# Designing robots with the context in mind- One design does not fit all

Ela Liberman-Pincu[1[0000-0002-9753-7714]], Elmer D. van Grondelle[2], and Tal Oron-Gilad[1[0000-0002-9523-0161]]

[1] 1 Ben-Gurion University of the Negev, Beer-Sheva, Israel
[2] Delft University of Technology, Delft, The Netherlands
elapin@post.bgu.ac.il, e.d.vangrondelle@tudelft.nl, orontal@bgu.ac.il

**Abstract.** Robots' visual qualities (VQs) impact people's perception of their characteristics and affect users' behaviors and attitudes toward the robot. Recent years point toward a growing need for Socially Assistive Robots (SARs) in various contexts and functions, interacting with various users. Since SAR types have functional differences, the user experience must vary by the context of use, functionality, user characteristics, and environmental conditions. Still, SAR manufacturers often design and deploy the same robotic embodiment for diverse contexts. We argue that the visual design of SARs requires a more scientific approach considering their multiple evolving roles in future society. In this work, we define four contextual layers: the domain in which the SAR exists, the physical environment, its intended users, and the robot's role. Via an online questionnaire, we collected potential users' expectations regarding the desired characteristics and visual qualities of four different SARs: a service robot for an assisted living/retirement residence facility, a medical assistant robot for a hospital environment, a COVID-19 officer robot, and a personal assistant robot for domestic use. Results indicated that users' expectations differ regarding the robot's desired characteristics and the anticipated visual qualities for each context and use case.

**Keywords:** context-driven design, visual qualities, socially assistive robot.

## 1 Introduction

### 1.1 Socially Assistive Robots

Recent years indicate a growing need for Socially Assistive Robots (SARs) [1-3]. Examples of SARs exist in the para-medical field [4,5] and for domestic use, taking care of the elderly or people with disabilities [6,7], or helping with children [8,9]. Since SAR types have functional differences, we expect the user-robot interaction to vary by use context, functionality, user characteristics, and environmental conditions [10,11]. Yet, our market research revealed that SAR manufacturers often design and deploy the same robotic embodiment for diverse contexts (see section 1.3, Table 2). Most studies in the field of SARs' appearance evaluate users' perceptions of existing off-the-shelf

SARs [12-15]. Few studies looked at isolated visual qualities using designated SARs [16].

The lack of design research, standards, or a consistent body of knowledge in this field forces designers to start from scratch when designing new robots [17,18]. Thus, the design of SARs requires a more scientific approach considering their evolving roles in future society. Technological products, even innovative and cutting-edge ones, often fail in the market when the design does not evoke the desired human cognitive response and action, does not apply to environmental conditions, or leads to unrealistic expectations [19-21]. When developing new SARs, the focus is mainly on guaranteeing functionality and safety. Aesthetics and robot look are part of the design process but not necessarily context-specific, i.e., one design fits all. Hekkert and van Dijk (2011) [22] suggest the designer should begin by defining a vision for the context and desired interaction of a new product to set the most appropriate visual qualities (VQs) and come up with a suitable solution for particular design problems.

In this paper, we first classify SARs by their use contexts by outlining four contextual layers: the domain in which the SAR exists, its physical environment, intended users, and role. For example, a robot for the para-medical field, supporting non-professional older adults in their private homes for physical exercises. Then, we examine how potential users perceive SARs in context by delving into the robots' essential characteristics and desired VQs. We used an online questionnaire to collect participants' expectations of robot characteristics by contexts of use and their related VQs. We then analyze this data to evaluate users' perceptions of each robot's desired character and the factors affecting the participant's selection of VQs. Finally, we compared the findings with previous work to form design tools to support user- and interaction-centered designs for diverse tasks and use cases of SARs.

### 1.2 Initial mapping of Visual Qualities Perceptions

Previously we evaluated the effect of three VQs for SARs: body structure, outline, and color, on users' perception of the SAR's characteristics [23]. We have empirical findings on how isolated VQs impact people's perception of its characteristics: friendly, childish, innovative, threatening, old-fashioned, massive, elegant, medical, and the robot's gender, as presented in Table 1. For example, to achieve the perception of a friendly SAR, a designer should consider using A-shape or hourglass structure and avoid V-shape, choose light colors (e.g., a combination of white and blue), and avoid dark colors.

**Table 1.** VQs' effect on self-designed SAR characteristics. Dark boxes represent significant effects (adapted from Liberman Pincu et al. [23]).

| | Friendly | Childish | Innovative | Threatening | Old-fashioned | Massive | Elegant | Medical | Robot sex |
|---|---|---|---|---|---|---|---|---|---|
| Structure | ■ | | ■ | ■ | | ■ | ■ | | ■ |
| Color | ■ | ■ | | ■ | | ■ | ■ | ■ | ■ |
| Outline | | | ■ | | ■ | | ■ | | |

Significance level p<.05

These links between SAR characteristics and VQs provide designers with an initial mapping for selecting the VQs most suitable to the robot's role and its desired characteristics, increasing the possibility of aligning with user expectations, at least for the initial encounters with the SAR.

## 2 Deconstruction of Contexts Layers- Domain, Environment, Users, and Role.

To further enhance the design guidelines to assist designers in the design process of a new SAR and to align these characteristics with the relationship models [24], we now evaluate user expectations in different contexts of use. We map the relevant and desired characteristics for different SARs by a deconstruction process parsing into four contextual layers: Domain, Environment, Users, and Role. The following sections details each layer.

### 2.1 Domains

Our literature survey and market research lead to seven popular domains for SARs: Healthcare (including Eldercare and Therapy), Educational, Authority (including Security), Companion, Home assistance, Business, and Entertainment. Table 2 provides examples for each domain.

**Table 2.** Common domains in the market and the literature

| Domain | Commercial examples | References |
|---|---|---|
| Healthcare | Temi (Robotemi), Pepper (Softbank), NAO (Softbank), Misty (Misty Robotics), QTrobot (LuxAI) | 2,4,5 |
| Educational | Pepper (Softbank), NAO (Softbank), QTrobot (LuxAI), Buddy (Blue Frog Robotics) | 25-27 |
| Authority | Knightscope (Knightscope), Cobalt (cobalt robotics) | 28-31 |
| Companion | Temi (Robotemi), Misty (Misty Robotics), Buddy (Blue Frog Robotics), Aido (Aido) | 32-34 |
| Home assistance | Misty (Misty Robotics), Aido (Aido) | 6,35,36 |
| Business | Temi (Robotemi), Pepper (Softbank), NAO (Softbank), Cobalt (cobalt robotics), Buddy (Blue Frog Robotics) | 37,38 |
| Entertainment | Pepper (Softbank), NAO (Softbank), Buddy (Blue Frog Robotics), Aido (Aido) | 39,40 |

### 2.2 Physical Environments

SARs are intended for varied environments. The basic level refers to the robot's intended physical location: indoor or outdoor. This classification affects many engineering decisions considering environmental conditions such as light, noise, humidity, dust, and surface conditions (floor, carpet, grass, etc.). The second level refers to privacy: Personal (i.e., home or private office); Semi-public, meaning there are different users all familiar with the robot (e.g., workplace, assisted living residence, etc.); or public, meaning there are multiple users, some are passersby interacting with the robot for the first time. Figure 1 illustrates the levels of physical environments.

### 2.3 Users

Users can be classified by demographic information, like gender, age, or culture, or by their needs, abilities, and disabilities (cognitive and physical) [41,42]. In addition, users can be professional (trained to work with the robot, e.g., a trained nurse working with a medical robot) or non-professional [43] (e.g., a hotel guest interacting with a receptionist robot), as well as random (occasional passersby) or familiar with the robot (regularly interacting in the workplace or elsewhere but not as part of their professional work, e.g., a security robot placed at a building entrance). Figure 2 illustrates a classification of users.

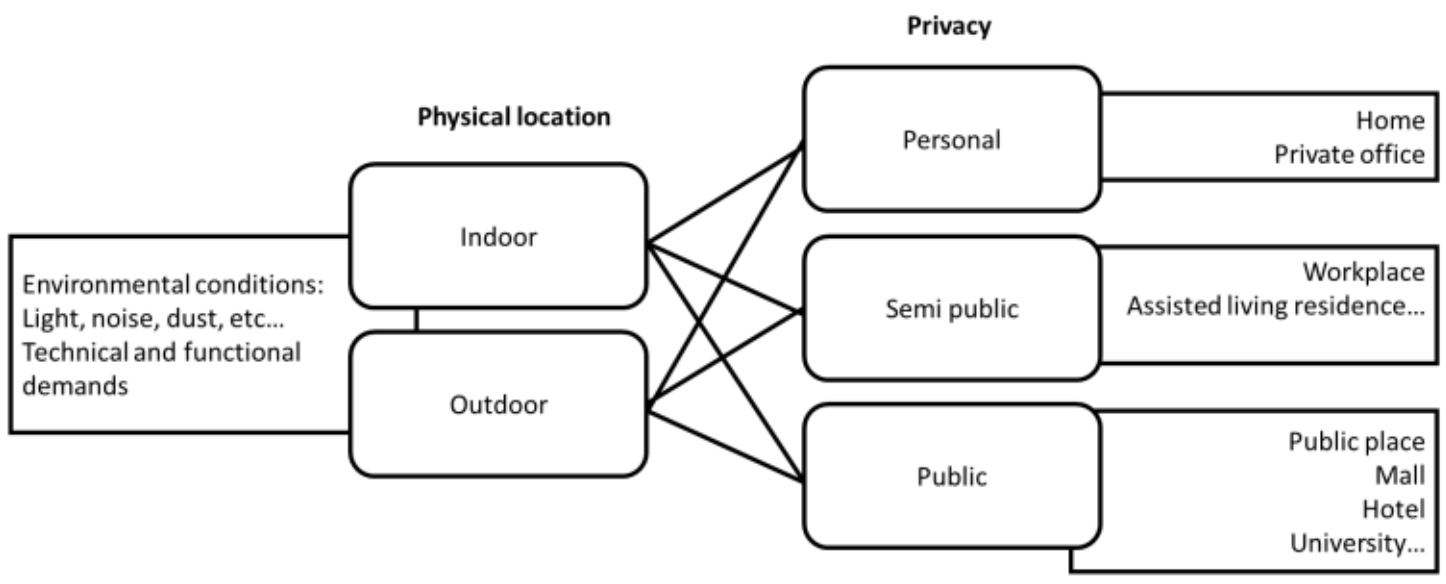

**Fig. 1.** The two levels of physical environments (physical location and privacy).

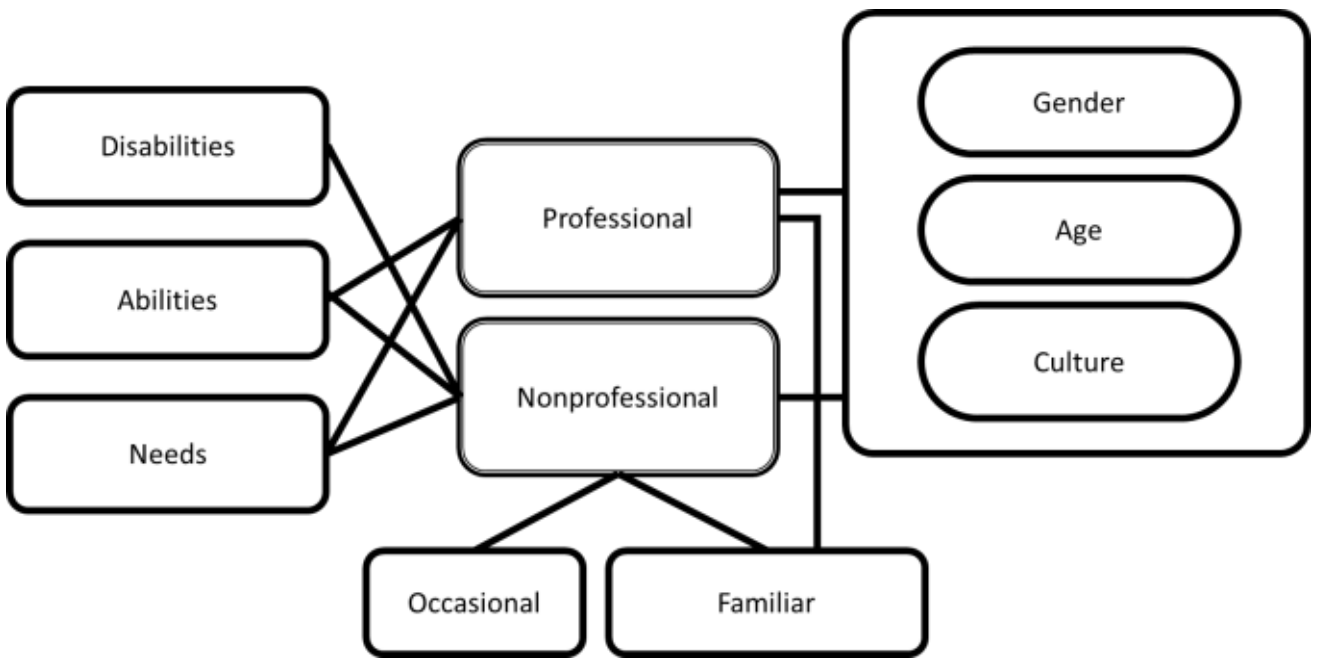

**Fig. 2**. Classification of users by their demographics, characteristics, and familiarity with the robot.

### 2.4 Roles and tasks

The human-robot relationship links to the robot's role and tasks, answering questions such as: Is this robot here to help me? In what way? Different human-robot relationship theories suggest classifying relationships by hierarchy [44]; Should I obey the robot? Who supervises who? Who leads the interaction?

Different role categorizations are found in the literature; Onnasch and Roesler (2021)[45], for example, classified eight abstract roles for various application domains: Information exchange, Precision (e.g., robots for micro-invasive surgery), Physical load reduction, Transport (transport objects from one place to another), Manipulation (the robot physically modifies its environment), Cognitive stimulation, Emotional stimulation, Physical stimulation. Abdi et al. [46] identified five roles of SAR in elder care: affective therapy, cognitive training, social facilitator, companionship, and physiological therapy.

For our model, we used the eight roles based on Onnasch and Roesler (2021)[45], excluding *precision* (which is less related to social relationships) that was replaced with *regulation*: Information exchange, Physical load reduction, Transport, Manipulation, Cognitive stimulation, Emotional stimulation, Physical stimulation, and Regulation.

Each role is classified into a three-level hierarchy of human-robot relationships (robot-led interaction, equal or human-led interaction). For example, in the *physical stimulation* section, we can have a training robot at the *robot-led interaction* level, a teammate at the *equal* level, and a physical therapy robot at the *human-led interaction* level. Fig 3 illustrates the hierarchy of relationships.

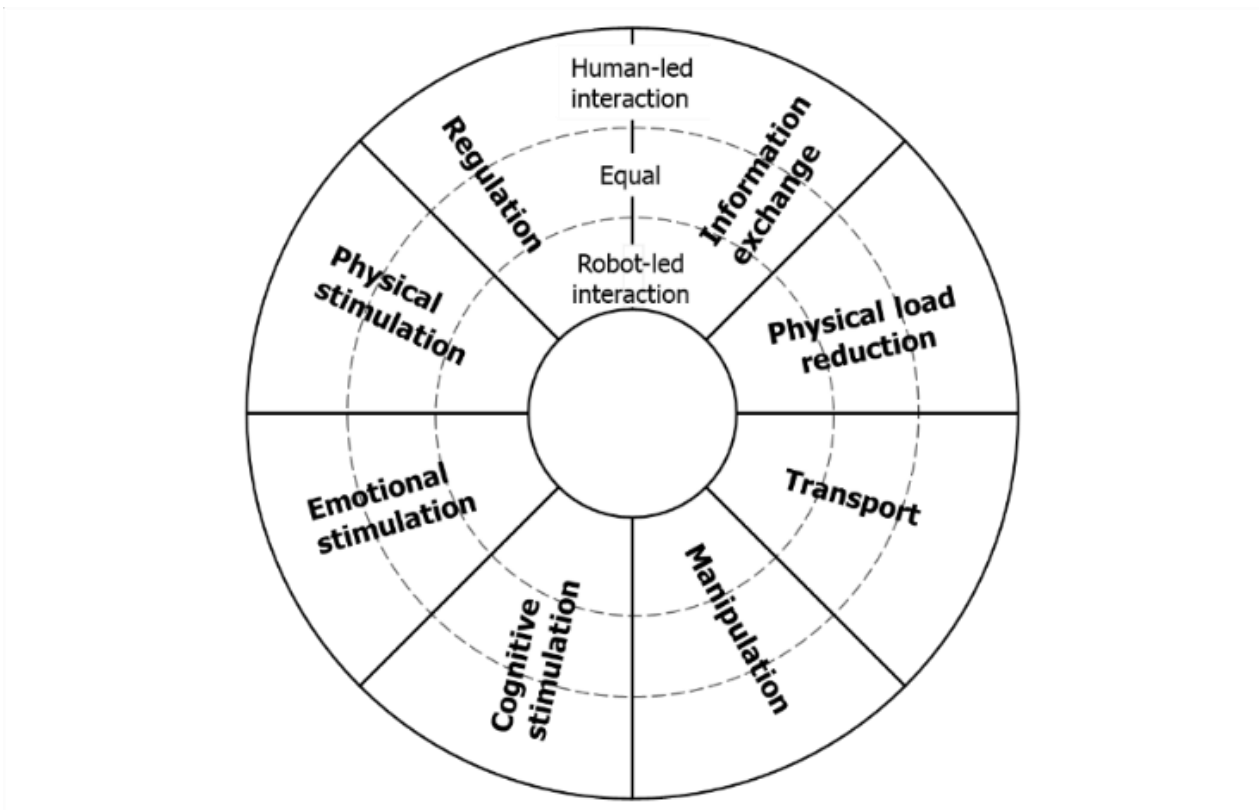

**Fig. 3.** Eight SAR roles (adapted from Onnasch and Roesler (2021)) by three hierarchy levels of leadership (robot-led interaction, equal or human-led interaction).

## 3 Evaluating Users' Expectations and design perceptions

### 3.1 Aim and Scope

Based on potential users' evaluations, this study aims to define the appropriate characteristics for robots in different contexts of use. The outcomes will be used together with previous findings [23,47] to form tools and guidelines for designers and manufacturers.

### 3.2 Four use cases

To apply the deconstruction layers of design, we defined four SAR use cases that differ by their contextual layers: a service robot for an Assisted Living/retirement residence facility (ALR), a Medical Assistant Robot (MAR) for a hospital environment, a Covid-19 Officer Robot (COR), and a Personal Assistant Robot (PAR) for home/domestic use. The following paragraph details each case. Table 3 summarizes them.

**Table 3.** Four SAR use cases

| | Domain | Environment | | Users | | Role |
|---|---|---|---|---|---|---|
| ALR | Business | An assisted living residence facility | Semi-public<br>Indoor | Older adults | Non-professional | Information exchange<br>Human-led interaction |
| MAR | Healthcare | Hospital | Public | Medical crews | Professional | Information exchange/ Transport |
| | | | Indoor | Hospitalized and caregivers | Non-professional | Human-led interaction / equal |
| COR | Authority | Public places | Public<br>Indoor/ outdoor | Passersby | Non-professional | Regulation<br>Robot-led interaction |
| PAR | Home assistance | Home | Personal<br>Indoor | Diverse | Non-professional | Physical load reduction/ Cognitive stimulation/ Emotional stimulation<br>Human-led interaction/ equal |

**A service robot for an Assisted Living/retirement residence facility (ALR)** aims to roam the lobby and be used by the facility residents to register for various classes and activities. In addition, it provides information and helps communicate (via video calls and chats) with staff members. **A Medical Assistant Robot (MAR) for a hospital environment** aims to assist the medical team, especially when social distancing is required. Through it, the medical team can communicate in video calls with isolated patients and bring equipment, food, and medicine into patients' rooms. **A COVID-19 Officer Robot (COR)** aims to ensure passersby comply with Covid-19 restrictions like social distancing or wearing a face mask. **A Personal Assistant Robot (PAR) for home/domestic use** seeks to assist users with daily tasks, recommend activities at home and outside, and remind them of their duties and appointments. The robot allows users to watch videos, listen to music, play, and have video chats with family and friends.

### 3.3 Evaluation Method and Online Questionnaire Design

Using Qualtrics, we designed an online questionnaire where participants were exposed to one of the four use cases. First, they were asked to define the robot's desired characteristics by marking relevant words out of a word bank. The word bank contained twelve words based on previous studies related to SARs' perception [48-50] and that were found relevant to our four use cases: innovative, inviting, cute, elegant, massive, friendly, authoritative, aggressive, reliable, professional, intelligent, and threatening. Following Benedek & Miner's product reaction cards (2002) [51], we followed a similar procedure to our previous studies; however, in this case, participants were not reacting to a design but an idea. In addition, they had the option of adding their own words. Following, they were asked to select three types of VQs: body structure, outline, and color scheme from a set of options, by the alternative that, in their opinion, best expresses the desired characteristics that they have chosen. Figure 4 illustrates the questionnaire design.

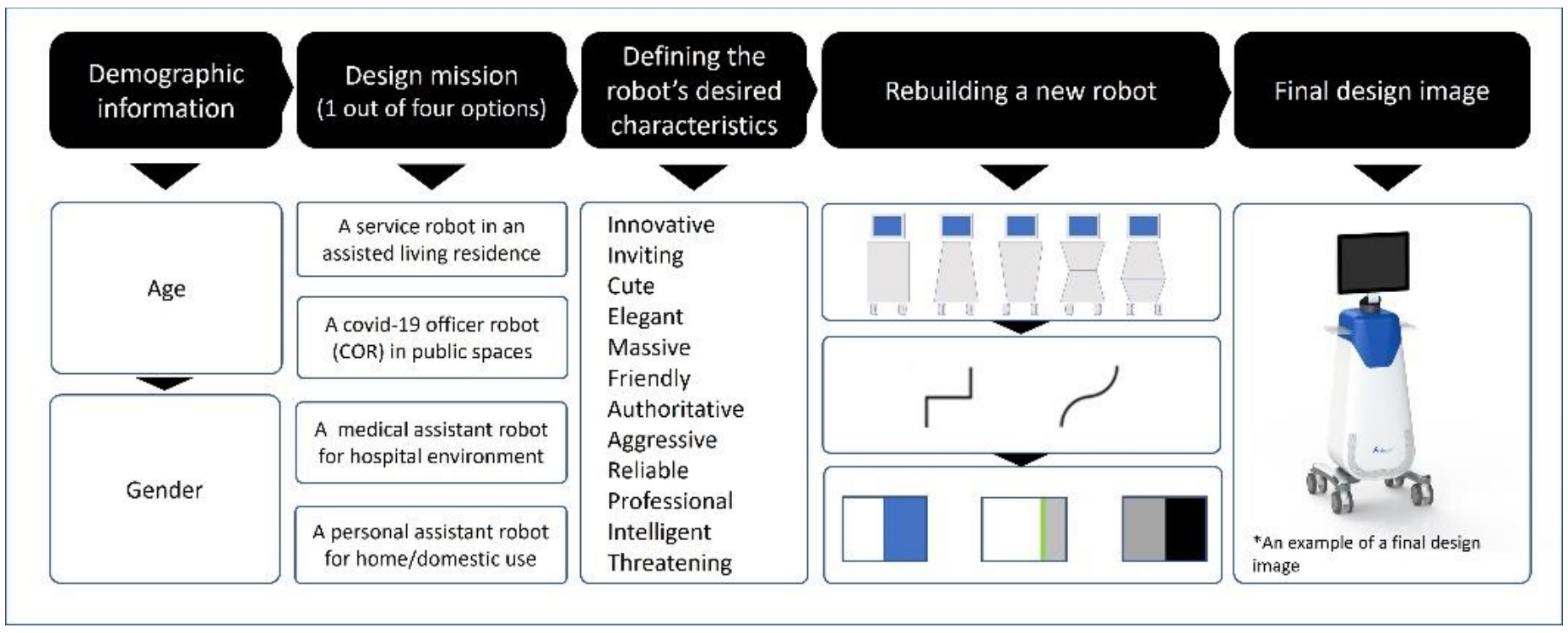

**Fig. 4.** questionnaire design.

The online questionnaire was distributed using social media and snowball distribution between November 2021 to March 2022 (via posts on Facebook and WhatsApp). In total, we collected data from 228 adult respondents. Table 4 Summarizes the respondents' demographics by use case.

**Table 4.** Summary of the respondents' data for each case study

| Case study | Gender | | | Age | Total |
|---|---|---|---|---|---|
| | Other | Males | Females | | |
| ALR | 1 | 29 | 24 | M=38.7, SD=17.5 | 54 |
| MAR | 1 | 24 | 23 | M=35.3, SD=12.2 | 48 |
| COR | - | 30 | 25 | M=37.7, SD=15.5 | 55 |
| PAR | 1 | 39 | 31 | M=43.0, SD=18.0 | 71 |
| **Total** | **3** | **122** | **103** | **M=39.0, SD=16.4** | **228** |

# 4 Results

## 4.1 The Effect of Context on Users' Expectations

We used Chi-square tests of independence to evaluate the effect of the context and the participants' demographic information on their selections of desired characteristics. Participants were grouped into three age groups: up to 29, ages 30-49, and 50 and above. Three participants preferred not to indicate gender; hence in our evaluations of gender effect, N=225 instead of N=228. The words *massive*, *aggressive*, and *threatening* were selected by less than 2% of the participants and therefore were excluded from our analysis.

Results confirm that users have different expectations regarding the robot's characteristics suitable for each context of use. Table 5 presents the top three words selected by participants for each context. Figure 5 illustrates the chosen words for each context in a radar chart.

We have found statistically significant relations between the contexts and four describing words: *Inviting, Friendly, Elegant,* and *Authoritative.* For example, 78% of the participants indicated that ALR should look inviting, much more than PAR (56%), MAR (54%), and COR (42%), $X^2$ (3, $N$ = 228) = 14.8742, $p$ < .01. The word *Friendly* is more likely to be ascribed to ALR (81%), MAR (81%), and PAR (75%) but less likely to be attributed to COR (55%), $X^2$ (3, $N$ = 228) = 13.1663, $p$ < .01. Elegant is more suitable for describing a PAR than all three other contexts, $X^2$ (3, $N$ = 228) = 11.5077, $p$ < .01. Finally, authoritative is significantly more suitable for describing a COR than all three other contexts, $X^2$ (3, $N$ = 228) = 44.4546, $p$ < .01.

**Table 5.** Top three words selected by participants for each context and their rate.

| Case study | Most selected word | 2nd word | 3rd word |
|---|---|---|---|
| ALR | Friendly (81%) | Inviting (78%) | Reliable (67%) |
| MAR | Friendly (81%) | Professional (67%) | Reliable (67%) |
| COR | Professional (67%) | Reliable (60%) | Authoritative (58%) |
| PAR | Friendly (75%) | Reliable (69%) | Professional (65%) |

In addition, we have found that the participants' demographic data (gender and age) affect their selections. Female participants were significantly more likely to select the words *Cute* ($X^2$ (1, $N$ = 225) = 6.68, $p$ < .01) and *Friendly* ($X^2$ (1, $N$ = 225) = 10.813, $p$ < .01). Though a Chi-square test of independence showed that there were no significant associations between gender and the selection of the word *Innovative,* male participants were more likely to select it (48%) than female participants (37%), ($X^2$ (1, $N$ = 225) = 2.503, $p$ = .11. The words *Aggressive* and *Threatening* were selected only by male participants.

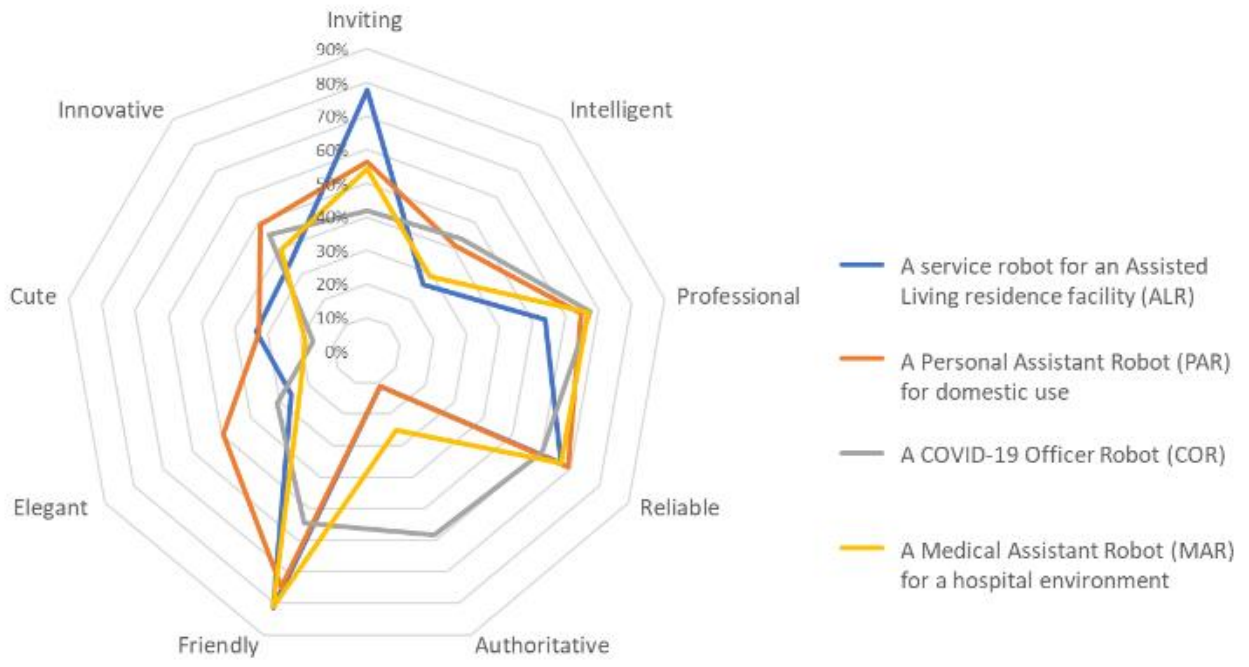

**Fig. 5.** Users' assigned characteristics by the context of use.

User age category was found to affect the desire for Innovative robots significantly; younger participants (up to 29) selected this character more frequently (53%) than mid-age (ages 30-49) (42%) and older (50 and above) participants (26%), the relationship between these variables was significant, $X^2$ (2, $N$ = 228) = 10.78, $p$ < .01. Furthermore, results indicate two worth noting yet, not statistically significant trends. Implying that age has a positive correlation with selecting the word *Cute* and a negative correlation with choosing the word Professional (i.e., older participants tend to desire cuter, less professional-looking robots). Table 6 summarizes the factors affecting the participants' selection of words.

**Table 6.** Factors affecting the participants' selection of words.

| | Innovative | Inviting | Cute | Elegant | Friendly | Authoritative |
|---|---|---|---|---|---|---|
| Context | | ■ | | ■ | ■ | ■ |
| Gender | | | ■ | | ■ | |
| Age | ■ | | | | | |

Significance level p<.05

### 4.2 Participants' selection of visual qualities

After selecting the characteristics of the SAR, participants were asked to choose the most suitable visual qualities for the context of use they had. Some VQs were selected more frequently than others regardless of the use context or other factors. For example, most respondents (79%) preferred rounded edges over chamfered ones. Only 10% of the respondents chose the dark color scheme, and most participants (49%) preferred the white color scheme. The two most selected structures were the Hourglass (27%) and the A shape (26%).

The context of use impacted just the body structure selection. For the ALR, respondents showed a higher preference for the A shape (37% compared to 26% in the overall data). PAR and COR increased the respondents' tendency to select the Hourglass structure (32% and 31%, respectively, compared to 27% in the overall data).

Participants' gender was found to affect their selection of colors significantly, $X^2$ (2, $N$ = 225) = 7.939, $p$ < .05; the male participants were more likely to select the white and blue combination (54%), while the female participants preferred the white option. Participants' age did not affect their selections, although there were minor differences among the three groups.

Participants' expectations (according to the selected words) significantly affected their selection of several VQs. Wanting to express *Inviting* increased the participants' probability of selecting a rounded outline, $X^2$ (1, $N$ = 228) = 7.946, $p$ < .01, and decreased the participants' probability of selecting the *Dark* color scheme, $X^2$ (2, $N$ = 228) = 5.537, $p$ = .06. To express *Cute* participants selected using the *white* color, $X^2$ (2, $N$ = 228) = 16.03, $p$ < .01, and were more likely to select the *A shape* structure, $X^2$ (4, $N$ = 228) = 13.23, $p$ < .05. Participants who wished to express *Elegant* showed a tendency to select chamfered outline significantly more than in the general population of the study (36% compared to 21% in the overall data), $X^2$ (1, $N$ = 228) = 23.26, $p$ < .01. Wanting to express *Friendly* increased the participants probability to select the *Hourglass* structure, $X^2$ (4, $N$ = 228) = 9.5, $p$ < .05, the *Rounded* outline $X^2$ (1, $N$ = 228) = 6.43, $p$ < .05, and the *White and blue* color combination, $X^2$ (2, N = 228) = 8.15, p < .05. Table 7 concludes our findings.

**Table 7.** Participants' selection of Visual qualities. Gray boxes represent a significant level of p<.05; black boxes represent a significance level of p<.01.

| | Structure<br>(Five levels) | Outline<br>(Two levels) | Color<br>(Three levels) |
|---|---|---|---|
| Overall | A shape (26%)<br>Diamond (18%)<br>Hourglass (27%)<br>Rectangle (10%)<br>V shape (19%) | Rounded (79%)<br>Chamfered (21%) | Dark (10%)<br>White (49%)<br>White and blue (41%) |
| Assisted Living service robot (ALR) | A Shape (37%) | Rounded (78%) | White (46%) |
| Personal assistant robot (PAR) | Hourglass (32%) | Rounded (85%) | White (46%) |
| COVID-19 Officer Robot (COR) | Hourglass (31%) | Rounded (75%) | White (47%) |
| Medical Assistant Robot (MAR) | - | Rounded (77%) | White (56%) |
| Male | - | Rounded (79%) | White and blue (54%) |
| Female | - | Rounded (80%) | White (50%) |
| Up to the age of 29 | Hourglass (31%) | Rounded (80%) | White and blue (54%) |
| Ages 30-49 | - | Rounded (75%) | White and blue (48%) |
| 50 and above | A Shape (30%) | Rounded (83%) | White (53%) |
| Inviting | - | Rounded** | Not Dark |
| Cute | A shape* | - | White** |
| Elegant | - | ** | - |
| Friendly | Hourglass* | Rounded * | White & blue* |

# 5 Discussion and Future Work

SARs are becoming more prevalent in everyday life [1-3], establishing different kinds of relationships [44]. The body of knowledge in human-robot interaction keeps growing to ensure that these robots follow human' social norms and expectations [10,11,52]. However, knowledge is limited regarding the design research of SARs [17-18]. Most research in the field focuses on evaluating users' perceptions of existing off-the-shelf SARs [12-15]. In this work, we sought to define what these expectations are. Using an online questionnaire, we collected data from 228 respondents regarding their expectations from SARs in four use cases differ in their four layers of context. Results confirm that users have different expectations regarding the robot's characteristics that are suitable for each context. However, in most cases (excluding COR), the top selected word was *Friendly* (see section 4.1 and Table 5).

Further, when asked to select VQs that best express these expectations, we found that participants' demographic data significantly affected their selections (see section 4.2, and Table 7). The four final designs (formed by looking at most users' selections) are almost similar and differ only by the structure; all designs are rounded-edged white robots. Fig 6 presents the four designed robots by use case.

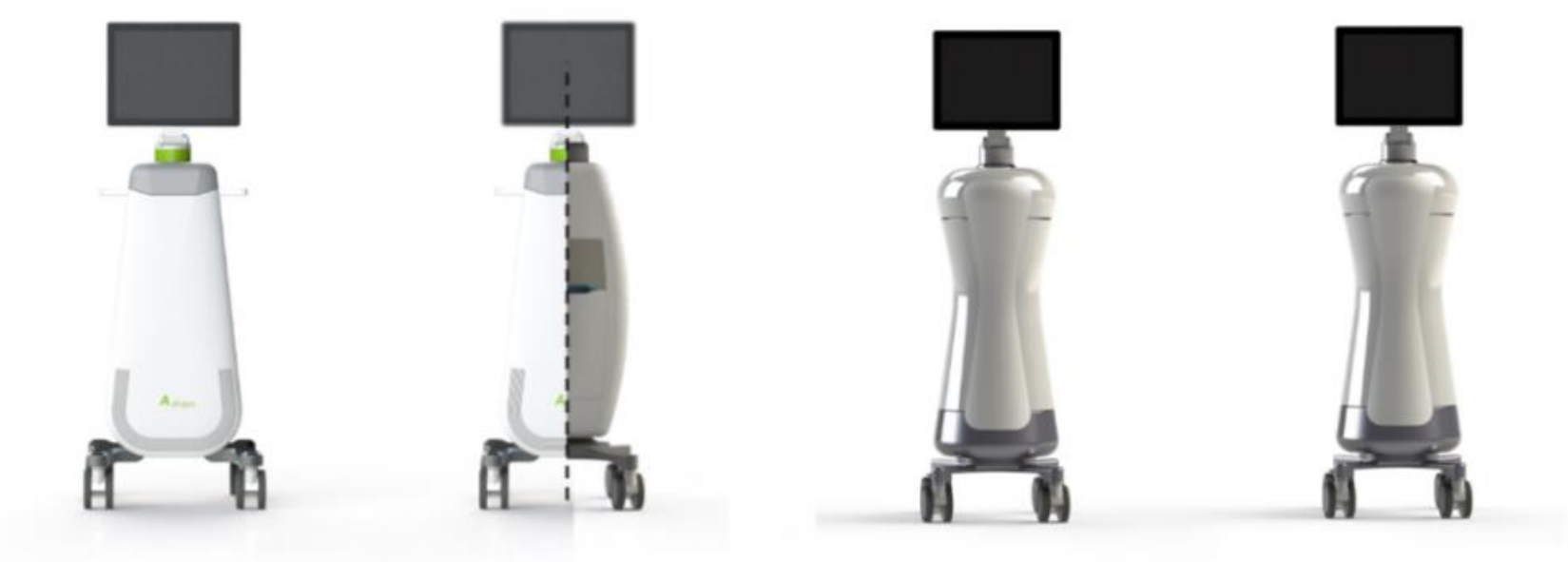

| a service robot for an Assisted Living/retirement residence facility (ALR) | Medical Assistant Robot (MAR) for a hospital environment | a COVID-19 Officer Robot (COR) | a Personal Assistant Robot (PAR) for domestic use |
|---|---|---|---|
| A Shape (37%)<br>Rounded (78%)<br>White (46%) | A Shape (27%)/Diamond (25%)<br>Rounded (77%)<br>White (56%) | Hourglass (31%)<br>Rounded (75%)<br>White (47%) | Hourglass (32%)<br>Rounded (85%)<br>White (46%) |

**Fig. 6.** Four options of robots design by use case (by looking at the majority of users' selections). See Table 7 for the full details.

We then combined our findings of this study with our previous empirical findings [23] to set up design guidelines to assist designers in creating a new SAR according to its context. Figure 7 presents desired characteristics in each context according to our recent findings; the table on the right presents design suggestions for each characteristic. For example, in designing a new service robot for an assisted living residence facility (ALR), designers should inspire for a friendly, inviting, and reliable look; hence, they may consider choosing a rounded, white and blue, a-shaped design.

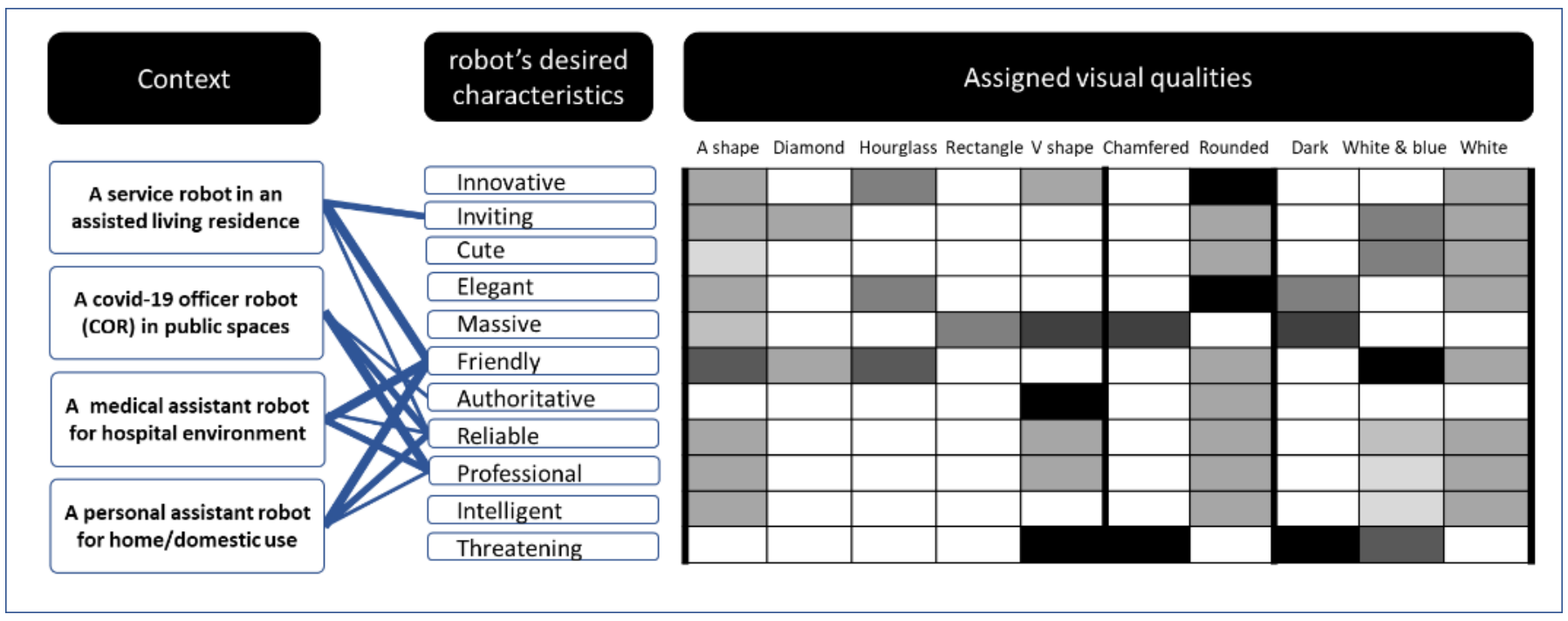

**Fig. 7** The desired characteristics in each context according to our recent findings; the table on the right presents design suggestions for each characteristic

The results align with previous studies that found that design preferences are more related to users' personal preferences. There is no consensus among **users** regarding the appropriate appearance for SARs. Hence, specifically for the case of a personal robot (PAR), the design should allow users to make adjustments using mass customization [53]. In addition, these results may indicate that the participatory design of SARs should be done carefully using two-way evaluations, allowing users to express themselves without relying on them to make design decisions. Participatory design outcomes should be assessed with other users using evaluation tools like Microsoft reaction cards [51].

This research, however, is subject to several limitations. First, participants were not asked to justify their selections; hence, we could not track their intentions. Second, the participants could only select three VQs: structure, outline, and color, with a closed set of options. Therefore, all outcomes share the exact proportions and dimensions. The robot's height affects user perception [54]; in our previous study evaluating the effect of the COVID-19 officer robot's appearance, some participants mentioned it should be taller [28]. Our subsequent studies will explore stakeholders' perceptions as well as the effect of culture. Such findings will provide further support for the design process of new SARs depending on their context of use, their intended role, and users. And form design guidelines for future SARs.

## 6 Acknowledgment

This research was supported by Ministry of Innovation, Science and Technology, Israel (grant 3-15625), and by Ben-Gurion University of the Negev through the Helmsley Charitable Trust, the Agricultural, Biological and Cognitive Robotics Initiative, the W. Gunther Plaut Chair in Manufacturing Engineering and by the George Shrut Chair in Human performance Management.